\documentclass[pdflatex,sn-vancouver,Numbered]{sn-jnl}

\usepackage{graphicx}%
\usepackage{multirow}%
\usepackage{amsmath,amssymb,amsfonts}%
\usepackage{amsthm}%
\usepackage{mathrsfs}%
\usepackage[title]{appendix}%
\usepackage{xcolor}%
\usepackage{textcomp}%
\usepackage{manyfoot}%
\usepackage{booktabs}%
\usepackage{algorithm}%
\usepackage{algorithmicx}%
\usepackage{algpseudocode}%
\usepackage{listings}%

\begin{document}

\title[Agentivism: a learning theory for the age of artificial intelligence]{Agentivism: a learning theory for the age of artificial intelligence}

\author*[1]{\fnm{Lixiang} \sur{Yan}}\email{lixiangyan@tsinghua.edu.cn}
\author*[2,3]{\fnm{Dragan} \sur{Gašević}}\email{dgasevic@hku.hk}

\affil[1]{\orgdiv{School of Education}, \orgname{Tsinghua University}}
\affil[2]{\orgdiv{Faculty of Education and School of Computing \& Data Science}, \orgname{The University of Hong Kong}}
\affil[3]{\orgdiv{Faculty of Information Technology}, \orgname{Monash University}}

\abstract{
Learning theories have historically changed when the conditions of learning evolved. Generative and agentic AI create a new condition by allowing learners to delegate explanation, writing, problem solving, and other cognitive work to systems that can generate, recommend, and sometimes act on the learner's behalf. This creates a fundamental challenge for learning theory: successful performance can no longer be assumed to indicate learning. Learners may complete tasks effectively with AI support while developing less understanding, weaker judgment, and limited transferable capability. We argue that this problem is not fully captured by existing learning theories. Behaviourism, cognitivism, constructivism, and connectivism remain important, but they do not directly explain when AI-assisted performance becomes durable human capability. We propose Agentivism, a learning theory for human-AI interaction. Agentivism defines learning as durable growth in human capability through selective delegation to AI, epistemic monitoring and verification of AI contributions, reconstructive internalization of AI-assisted outputs, and transfer under reduced support. The importance of Agentivism lies in explaining how learning remains possible when intelligent delegation is easy and human-AI interaction is becoming a persistent and expanding part of human learning.
}


\maketitle

\section*{Main}

Learning theory has repeatedly changed when the conditions of learning changed. Behaviourism emerged when psychology sought lawful relations between environment and behaviour \cite{Watson1913,Skinner1938}. Cognitivism shifted attention to mental representation, memory limits, and control processes \cite{Miller1956,AtkinsonShiffrin1968}. Constructivist traditions reframed learning as active meaning making in experience and social interaction \cite{Dewey1938,Piaget1952,Vygotsky1978}. Connectivism responded to digital environments in which knowing increasingly depended on navigating distributed networks of information and expertise \cite{Siemens2005}. These traditions did not replace one another so much as respond to changes in what became psychologically salient about learning. Generative and agentic AI now create another such shift because they change not only access to knowledge, but the ease with which knowledge can be mobilized into task performance.

Generative and agentic AI make performance and learning more sharply separable. A learner no longer confronts only content, task demands, and human guidance, but may also interact with a system that can interpret prompts, synthesize prior cultural material, draft solutions, recommend next steps, and sometimes complete parts of the task itself \cite{Milano2023,Farrell2025,Collins2024,Extance2023}. Under these conditions, the central question is no longer only what the learner knows, but what the learner can later explain, justify, adapt, and transfer with reduced dependence on the same support. Emerging empirical research shows that this distinction is already consequential and measurable across productivity and learning outcomes \cite{NoyZhang2023,Ng2024SRLbotScience,Kavadella2024DentalChatGPT,Abdelhalim2025VocabularyChatGPT,DeSimone2025NigeriaTutoring}. In AI-assisted writing, learners can produce stronger essays while revising less independently and gaining no corresponding advantage in underlying knowledge, a pattern described as metacognitive laziness \cite{Fan2025,Playfoot2024}. In scientific inquiry, students using ChatGPT can report lower cognitive load while producing less sophisticated reasoning \cite{Stadler2024}. Related work likewise suggests that generative AI may improve measurable task performance without proportionate gains in metacognitive processing or independent understanding, making it necessary to distinguish performance gains from learning gains rather than treat them as interchangeable \cite{Fernandes2025,Yan2025Distinguishing,Li2025ProgrammingProcesses}.

This challenge is intensified by the socio-technical character of large AI models. Generative AI systems do not simply deliver information; they reorganize historically accumulated human artifacts through data regimes that privilege dominant languages, conventions, and frequencies \cite{Farrell2025,Brinkmann2023,LuSongZhang2025,Jin2025Metaphor}. Their outputs can therefore be rhetorically fluent, culturally patterned, and normatively averaged even when they appear individualized. Experimental evidence suggests that such systems can enhance individual or team creativity while in some cases reducing the diversity or semantic divergence of ideas produced collectively \cite{DoshiHauser2024,Xu2025CreativePotential,BaltaSalvador2026CreativeIdeation,Wei2025DigitalStoryGAI,Jin2026When,Sourati2026Homogenizing}. At the same time, low-friction support can invite overconfidence, cognitive offloading, and illusions of understanding \cite{MesseriCrockett2024,Clark2025,Wei2025DigitalStoryGAI,Cheng2026sycophantic}. The theoretical problem is therefore not simply whether AI helps learners complete tasks, but how learners exercise judgment, preserve agency, and develop durable capability when assistance is persuasive, fluent, and easy to accept.

Agentivism is proposed as a learning theory for this new condition. Agentivism is a mid-range conceptual theory \cite{Merton1968} of learning through human-AI interaction. It defines learning as durable growth in human capability that occurs when learners delegate selectively to AI, monitor and verify AI contributions, and reconstruct AI-assisted performance into knowledge and skill that remain available beyond the immediate interaction. Agentivism does not replace behaviourism, cognitivism, constructivism, or connectivism; rather, it reorganizes their insights around a new problem: when intelligent delegation becomes easy, learning can no longer be inferred from performance alone. This paper is therefore concerned primarily with learning rather than with educational systems in the broadest sense. Pedagogical design, assessment, governance, and inclusion matter here as conditions shaping learning processes, but the central question remains straightforward: when AI can contribute directly to task completion, what must happen for the learner, rather than only the system, to become more capable?

\begin{center}
\fbox{
\parbox{0.94\textwidth}{
\textbf{Box 1. Key terminology}

\textbf{Generative AI.} AI systems that generate new content such as text, images, code, audio, or other symbolic outputs in response to prompts or other inputs. In learning contexts, generative AI can provide explanations, drafts, summaries, examples, feedback, and suggested solutions.

\textbf{Agentic AI.} AI systems that do more than generate content by also initiating, sequencing, or executing task-relevant actions toward a goal. In learning contexts, agentic AI may decompose tasks, recommend next steps, retrieve resources, coordinate subtasks, or act with partial autonomy within learner-defined or system-defined constraints.

\textbf{Human-AI interaction for learning.} Task-oriented interaction in which a learner engages with a generative or agentic AI system while pursuing understanding, problem solving, writing, inquiry, or other learning-relevant activity.

\textbf{Assisted performance.} Successful task completion achieved with AI support, without sufficient evidence that the learner has developed durable understanding or transferable capability.

\textbf{Durable human capability.} Capability that remains available to the learner beyond the immediate interaction and can be explained, adapted, or transferred with reduced dependence on the same AI support.

\textbf{Delegation.} The allocation of part of a task to an AI system, including idea generation, drafting, summarizing, suggesting, or other task-relevant operations.

\textbf{Verification.} The learner's evaluation of AI-generated outputs in relation to evidence, reasoning, task requirements, and alternative interpretations.

\textbf{Reconstructive internalization.} The process by which a learner reworks AI-assisted outputs into knowledge or skill that can later be used independently or with less support.

\textbf{Transfer under reduced support.} The criterion that learning has occurred when the learner can later explain, adapt, or apply what was previously achieved with AI support under conditions of less or no support.
}}
\end{center}

\section*{Classical theories still matter, but stop short}

Behaviourism remains important because learning through human-AI interaction is still shaped by reinforcement, feedback, effort reduction, and repetition \cite{Watson1913,Skinner1938}. Generative AI systems are attractive in part because they provide immediate reward in the form of speed, fluency, correctness cues, and reduced effort. This helps explain why learners may quickly develop habits of relying on AI support: the interaction is often efficient, responsive, and subjectively satisfying. Empirical research reflects exactly these conditions. Across human-AI interaction tasks, learners can experience lower perceived burden and greater ease, precisely the kinds of conditions under which delegation can become reinforcing \cite{Ngu2025Game,Pan2025,Yan2025LLMCollabProgramming,Song2025DialogicDynamics,Fan2025AIPairProgramming}. Yet behaviourism alone cannot explain whether the relevant competence remains in the learner once support is removed. It can explain why AI use becomes habitual; it cannot determine whether repeated success reflects learning or merely successful outsourcing. Nor can it adequately explain why designs that require explanation or justification before accepting AI advice reduce over-reliance even when they make the interaction feel less convenient \cite{Bucinca2021}. In short, behaviourism explains why AI support can become behaviourally compelling, but not when such reinforcement produces durable human capability rather than dependence.

Cognitivism remains indispensable because memory limits, schema construction, retrieval, cognitive load, and control processes still constrain human learning with AI, just as they constrain learning without it \cite{Miller1956,AtkinsonShiffrin1968}. Indeed, the rise of generative AI makes some cognitivist concerns even more salient by making cognitive offloading easier and more attractive. Learners can use AI to summarize, draft, organize, or suggest without necessarily constructing the internal representations needed for later explanation and transfer. Empirical findings already point to this tension. Early research established that feeling of learning and actual learning can diverge sharply, even in active instruction contexts \cite{Deslauriers2019}; this divergence becomes more pronounced when external systems handle the cognitive work. Students using ChatGPT in inquiry tasks may report lower cognitive load while producing less sophisticated reasoning \cite{Stadler2024}, and learners may achieve stronger measurable performance with AI support without corresponding gains in metacognitive processing or retained understanding \cite{Bastani2025,Liu2025,Li2025ProgrammingProcesses}. Cognitivism therefore helps explain why offloading can alter mental processing, but it does not by itself specify when offloading remains educationally productive and when it becomes substitutive. In particular, it does not yet provide a sufficient account of how AI-assisted outputs are converted back into durable knowledge and skill that the learner can later use with reduced support.

Constructivist traditions remain essential because learning is still a matter of meaning making, interpretation, participation, and identity formation in socially organized activity \cite{Dewey1938,Piaget1952,Vygotsky1978}. This matters even more when AI enters writing, inquiry, explanation, and problem solving in conversational form \cite{Song2025DialogicDynamics,Hu2025CollaBot,Hu2025OCWAgent,Guan2025TripartiteEFL,Xiao2025DialogicReading}. Learners do not merely retrieve information from a passive tool; they respond to suggestions, negotiate wording, evaluate alternatives, and position themselves in relation to machine-generated contributions. Studies of learners revising AI-generated writing show substantial variation in these orientations, ranging from compliance-oriented uptake to more transformative use that preserves voice and substantive ownership \cite{Kim2026,Jin2025Metaphor,Singh2024}. Studies of interaction with generative AI teachable agents likewise suggest that when greater authority is attributed to AI within the interaction, students may elaborate on task content while showing reduced initiative or altered participation patterns \cite{Xing2026}. Constructivism therefore remains vital for explaining why dialogue and participation matter, but it does not fully resolve the epistemic problem raised by generative AI: a system can occupy the interactional position of a knowledgeable other without satisfying the epistemic conditions that make such guidance trustworthy \cite{Salvi2025}. Conversational participation alone is therefore no guarantee of justified learning.

Connectivism remains highly relevant because knowledge is still distributed across networks of people, tools, and information resources \cite{Siemens2005}. That insight is foundational for understanding learning in digital environments and remains useful for understanding why learners increasingly rely on external systems rather than internal recall alone. However, generative and agentic AI alter what it means for knowledge to be distributed. In earlier networked environments, external nodes were often repositories, channels, or pathways through which learners navigated information. By contrast, generative AI systems can produce content, infer intent, recommend action, and reorganize the sequence through which a learner engages a task. The network now includes systems that do not merely store knowledge, but actively shape how knowledge is represented, prioritized, and mobilized in the moment of use \cite{Wang2023}. This is visible not only for students but also for teachers and professionals, whose interaction with generative AI can reshape judgment, role distribution, and the organization of collective practice \cite{Tan2026}. Connectivism thus explains why learning increasingly depends on distributed access, but it does not fully explain what happens when some nodes in that network become generative, persuasive, and partially agentic. Distribution remains necessary as a description, but it is no longer sufficient as an explanation of how durable human learning unfolds.

Prior learning theories explain important parts of learning with AI, but none fully explains learning under conditions of intelligent delegation. Behaviourism explains reinforcement, cognitivism explains cognitive offloading and mental processing, constructivism explains interaction and meaning making, and connectivism explains distribution across networks. What remains insufficiently explained is the central issue now posed by generative and agentic AI: when part of task completion can be delegated to an intelligent system, under what conditions does performance become durable human capability rather than merely successful assisted performance?

\begin{table*}[t]
\centering
\footnotesize
\caption{What classical learning theories can explain, and where they fall short in the age of generative and agentic AI}
\label{tab:classical_theories_ai}
\renewcommand{\arraystretch}{1.2}
\begin{tabular}{p{2cm} p{6.2cm} p{6.2cm}}
\toprule
\textbf{Theory} & \textbf{Can explain} & \textbf{Falls short} \\
\midrule
\textbf{Behaviourism} 
& Why AI use can become reinforcing through immediate feedback, fluency, speed, effort reduction, and repetition \cite{Watson1913,Skinner1938,Ngu2025Game,Pan2025}. 
& Whether repeated AI-supported success reflects durable learning or merely reinforced dependence; why slowing users down through justification can reduce over-reliance \cite{Bucinca2021}. \\
\midrule
\textbf{Cognitivism} 
& How AI changes cognitive load, memory demands, schema construction, retrieval, and offloading during learning \cite{Miller1956,AtkinsonShiffrin1968,Stadler2024,Bastani2025,Liu2025}. 
& When offloading supports learning versus substitutes for it; how AI-assisted performance becomes retained, transferable human competence. \\
\midrule
\textbf{Constructivism} 
& Why dialogue, interpretation, participation, and meaning making remain central when learners interact with AI conversationally \cite{Dewey1938,Piaget1952,Vygotsky1978,Kim2026,Jin2025Metaphor,Xing2026}. 
& Whether AI-generated guidance is epistemically trustworthy; interaction alone does not guarantee justified learning \cite{Salvi2025}. \\
\midrule
\textbf{Connectivism} 
& Why learning depends on distributed networks of people, tools, and information resources \cite{Siemens2005}. 
& What changes when network nodes become generative, persuasive, and partially agentic, shaping representation, priorities, and action in real time \cite{Wang2023,Tan2026}. \\
\bottomrule
\end{tabular}
\end{table*}

\section*{What generative and agentic AI changes}

Four developments make a new learning theory necessary.

Knowledge has become mobilizable. By mobilizable, we mean that generative AI can reorganize externally available knowledge into immediately usable explanations, plans, drafts, examples, and action sequences \cite{Singhal2023,Kraemer2025,Rao2025}. Search engines made information locatable; generative AI makes it rapidly usable in the moment of task performance. A model can turn a vague request into a literature summary, a lesson plan, a coding strategy, or a plausible explanation within seconds. The key learning variable is therefore no longer access alone, but what happens when external knowledge can be assembled into performance with minimal delay \cite{Clark2025,Farrell2025}. This transformation is not automatic or educationally neutral. Studies of AI-supported lesson design suggest that the quality of mobilization depends on learners' pedagogical understanding and prompt construction, indicating that human expertise still shapes whether rapidly assembled output becomes educationally meaningful \cite{Celik2025}. Work on interface and prompt design likewise shows that requiring learners to articulate goals, outline arguments, or compare AI output with source material can reduce blind uptake and promote more selective engagement \cite{Kim2026}. Mobilization therefore names a new condition of learning, not merely a new convenience: when knowledge can be converted quickly into usable output, learning depends on whether that conversion recruits human judgment and reconstruction or bypasses them.

Agency has become more dynamically allocated during learning. Earlier learning theories assumed that the learner remained the primary locus of task execution even when tools, teachers, or peers provided support. Generative and agentic AI complicate that assumption because learners can now delegate parts of planning, drafting, explanation, evaluation, and problem solving to systems that respond contingently and sometimes proactively. The issue is not whether learners possess agency in the abstract, but how agency is distributed across the interaction as the task unfolds. Bandura's distinction among direct, proxy, and collective agency remains highly relevant here \cite{Bandura2001}, but the proxy now takes a form that is unusually flexible, conversational, and adaptive. Empirical work suggests that learners differ markedly in how they manage this distribution. Some accept AI output with minimal resistance, some make only cosmetic revisions, and some rework suggestions extensively in service of their own purposes \cite{Darvishi2024,Zheng2025,Singh2024}. Other studies show that greater perceived AI authority can reduce learner initiative and reshape subsequent participation, indicating that agency is not merely a stable trait but an emergent property of interaction \cite{Xing2026,Jin2026When}. Learning theory must therefore explain not only self-regulation, but regulation of delegation: what learners keep, what they offload, and what they later reclaim.

Performance and learning have become more sharply separable. Generative AI allows learners to produce polished outputs without commensurate growth in understanding, reasoning quality, or later independent capability. This possibility has always existed in some form, but AI scales it, accelerates it, and normalizes it across everyday tasks \cite{Eloundou2024}. The empirical literature increasingly converges on this concern. Learners can produce stronger written products with AI support while gaining no corresponding advantage in underlying knowledge and engaging in less independent revision \cite{Fan2025}. Students using ChatGPT in inquiry tasks can feel cognitively supported while producing less sophisticated reasoning \cite{Stadler2024}. Related studies likewise suggest that AI may improve measurable task performance without proportionate gains in metacognitive processing, retained understanding, or independent transfer \cite{Fernandes2025,Bastani2025,Li2025ProgrammingProcesses}. This is the point at which Agentivism departs most clearly from performance-centred accounts. Under conditions of intelligent delegation, a correct answer, a fluent essay, or an efficient workflow is no longer sufficient evidence that learning has occurred. Learning must instead be judged by what the learner can later explain, adapt, and transfer with reduced dependence on the same support.

Epistemic trust and diversity have moved from the periphery to the centre of learning. Large AI models are trained on historically accumulated and unevenly distributed human outputs, and their responses reflect the dominant patterns, omissions, and cultural tendencies of those corpora \cite{Brinkmann2023,LuSongZhang2025}. As a result, they do not merely support cognition; they also shape what appears plausible, salient, and worth saying. Experimental evidence suggests that generative AI can enhance individual creativity while reducing collective diversity of ideas \cite{DoshiHauser2024}, and biased AI writing assistance can influence individuals' attitudes and judgements rather than merely help them express pre-existing views \cite{WilliamsCeci2026}. At the same time, fluent and persuasive outputs can invite overconfidence, cognitive offloading, and illusions of understanding \cite{MesseriCrockett2024,Clark2025}. Learning under these conditions therefore requires more than access to useful output. It requires calibration of trust, attention to provenance and evidence, and vigilance about whose knowledge has been averaged, whose perspective has been marginalised, and how repeated reliance on the same systems may narrow inquiry over time \cite{Traberg2026,Hao2026,Sourati2026Homogenizing}. In this sense, epistemic judgement is no longer peripheral to learning with AI; it becomes part of the mechanism by which durable human capability is either strengthened or hollowed out.

\section*{Agentivism}

\textit{Agentivism} is a mid-range learning theory for human-AI interaction. It defines learning as durable growth in human capability that occurs when learners delegate selectively to generative or agentic AI, monitor and verify AI contributions, and reconstruct AI-assisted performance into knowledge and skill that remain available beyond the immediate interaction. The theory begins from a simple premise: when AI systems can contribute directly to task completion, learning can no longer be inferred from performance alone.

Two commitments distinguish Agentivism from more generic accounts of AI use in learning. First, Agentivism treats the allocation of agency during human-AI interaction as a central explanatory variable. A task is no longer carried entirely by the learner, nor merely supported by a passive tool. Instead, parts of planning, drafting, explanation, evaluation, and revision may be distributed across learner and system in ways that change what the learner actually practices and retains. The central theoretical question is therefore not whether AI is present, but how responsibility for cognitive and epistemic activity is allocated as the task unfolds. Second, Agentivism distinguishes assisted performance from learning. A learner has learned only if capabilities supported during interaction can later be explained, adapted, and transferred with reduced dependence on the same support. Verification, judgment, and reconstruction are therefore not auxiliary concerns added for responsible use; they are part of the learning process itself.

\begin{figure}
    \centering
    \includegraphics[width=\linewidth]{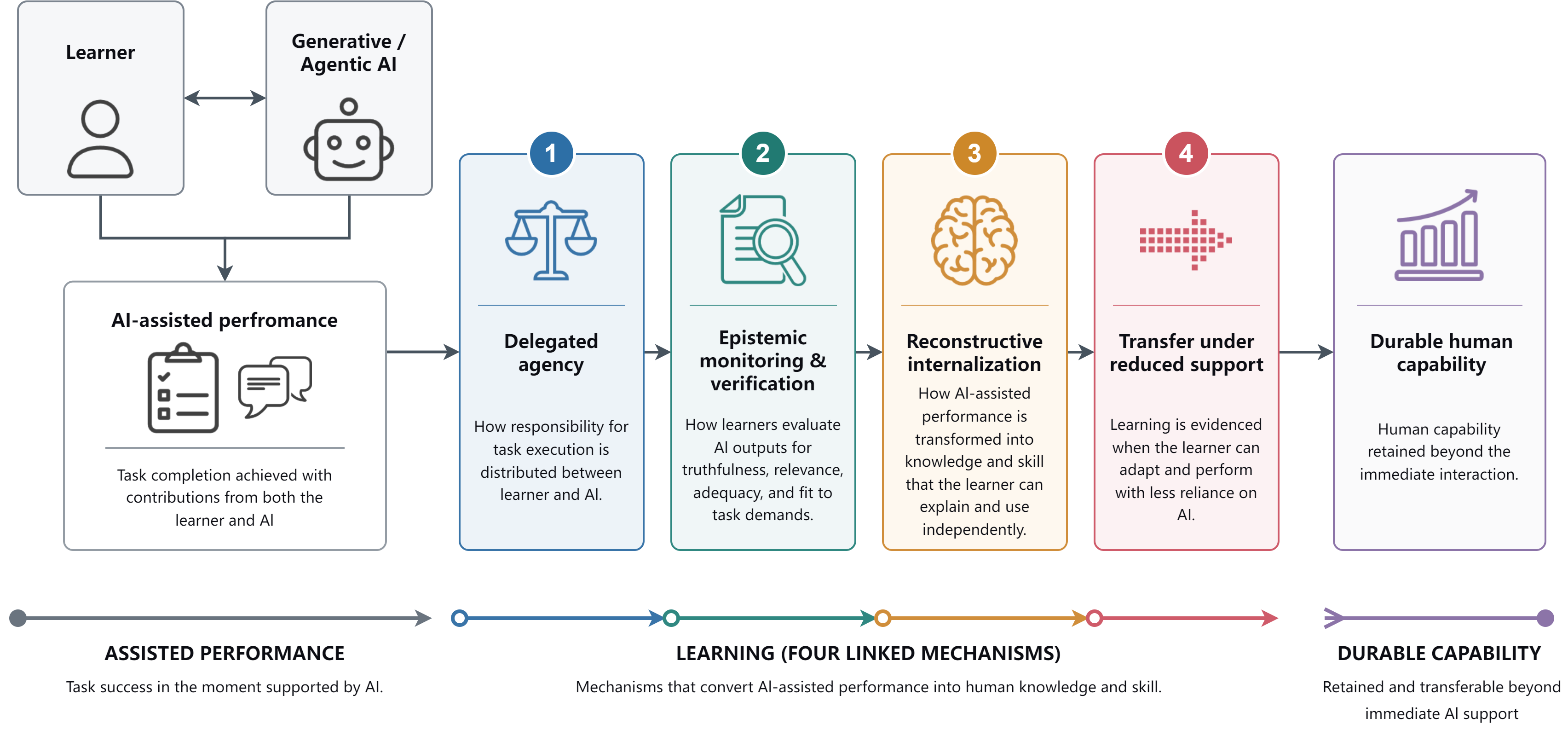}
    \caption{\textbf{Conceptual model of Agentivism as a learning theory for human-AI interaction.} Agentivism explains learning as durable growth in human capability under conditions where generative or agentic AI contributes directly to task completion. The figure presents four linked mechanisms. Delegated agency describes how responsibility for task execution is distributed between learner and AI. Epistemic monitoring and verification captures how learners evaluate AI outputs for truthfulness, relevance, adequacy, and fit to task demands. Reconstructive internalization explains how AI-assisted performance is transformed into knowledge and skill that the learner can explain and use independently. Transfer under reduced support indicates that learning is evidenced when the learner can adapt and perform with less reliance on AI. Together these mechanisms distinguish assisted performance from learning and specify when human capability is retained beyond the immediate interaction.}
    \label{fig:agentivism}
\end{figure}

Agentivism does not reject classical theories. What Agentivism adds is a reorganization of their insights around a new learning condition: when intelligent delegation becomes easy, the central issue is how learners remain the locus of durable capability even when parts of task performance are distributed across human and artificial contributors. For this reason, Agentivism should be understood neither as a full theory of educational systems nor as a narrow design framework. It is a conceptual learning theory aimed at explaining how learning unfolds when generative and agentic AI can contribute substantively to task completion. Pedagogical arrangements, interface designs, assessment regimes, and institutional rules matter in this account, but they matter mainly as conditions that shape learning processes rather than as substitutes for theorizing those processes. The theory's explanatory core lies in four linked mechanisms (Fig.~\ref{fig:agentivism}): delegated agency, epistemic monitoring and verification, reconstructive internalization, and transfer under reduced support.

\section*{Core mechanisms of learning under Agentivism}

\subsection*{Delegated agency}
Learning under Agentivism begins with delegated agency. In many forms of human-AI interaction, learners no longer perform every part of a task themselves. They may ask AI to generate options, draft text, summarize sources, propose explanations, or suggest next steps. What matters for learning is therefore not simply whether AI is used, but how responsibility for task execution is distributed across learner and AI system as the activity unfolds. Delegation can preserve learning when the learner remains responsible for framing the problem, setting criteria, and deciding what counts as acceptable reasoning or evidence. Delegation undermines learning when these functions silently migrate to the system and the learner becomes mainly a selector or acceptor of fluent outputs. Empirical studies already illustrate this variation: in online collaborative learning, arrangements that positioned AI as feedforward and feedback support versus AI partner produced different patterns of cognitive engagement and regulation, with the strongest outcomes when human and AI contributions were explicitly coordinated \cite{Gyasi2025HumanAICollaboration}. Other work suggests that when greater authority is attributed to AI within the interaction, learner initiative and participation can shift accordingly \cite{Xing2026}. Delegated agency is therefore the first mechanism because it determines what kind of cognitive and epistemic work remains available for the learner to do.

\subsection*{Epistemic monitoring and verification}
Delegation alone does not produce learning, so the second mechanism is epistemic monitoring and verification. Because generative AI outputs are probabilistic, rhetorically fluent, and often persuasive, learners must evaluate them for truthfulness, relevance, adequacy, provenance, and fit to task demands. Under Agentivism, this checking is not an optional layer of responsible use added after cognition has already occurred. It is part of the mechanism by which learning is preserved in human-AI interaction. When learners inspect claims, compare alternatives, cross-check evidence, or justify why a suggestion should be accepted, they remain cognitively and epistemically engaged with the task \cite{Tzirides2024AILiteracy,Kavadella2024DentalChatGPT}. When they do not, fluent output can be mistaken for understanding, and interaction patterns that favour agreement over critique can further promote dependence and reduce independent judgment \cite{Cheng2026sycophantic}. Research on cognitive forcing interventions shows that requiring users to explain or justify AI-supported decisions can reduce over-reliance even when such designs feel less convenient \cite{Bucinca2021}. Related educational studies likewise suggest that learners may feel cognitively supported while engaging in shallower reasoning or weaker metacognitive processing \cite{Fernandes2025,Song2026,Li2025ProgrammingProcesses}. Epistemic monitoring and verification therefore explain how interaction with AI becomes either a site of judgment and learning or a pathway to passive acceptance.

\subsection*{Reconstructive internalization}
The third mechanism is reconstructive internalization. Learning occurs only when AI-assisted outputs are reworked into the learner's own explainable and usable capability. A learner may complete a task successfully with AI support, but learning has not yet occurred unless the learner can reconstruct why the accepted response is appropriate, identify when it would fail, adapt it to a new situation, or reproduce the underlying reasoning with less assistance. This mechanism is what converts assisted performance into retained capability. It also clarifies why revision, explanation, and re-description matter so much in AI-supported activity \cite{Chen2025UAVReflectiveScaffold,Makransky2025SenseMaking}. Studies of inquiry and reasoning with AI show that apparent success with AI can coexist with underdeveloped reasoning when learners rely on output without reconstructing the logic behind it \cite{Bastani2025,Qian2026}. In problem-solving and simulation environments, substantial reworking of AI-generated solutions is necessary for learning gains to materialize, showing that reconstruction is not domain-specific to writing \cite{Lim2025Simulation}. Reconstructive internalization therefore specifies the point at which external support becomes educationally productive: not when the system produces a usable answer, but when the learner turns that answer into personally available knowledge or skill.

\subsection*{Transfer under reduced support}
The fourth mechanism is transfer under reduced support. Agentivism treats later independent or less-supported performance as the criterion by which learning is distinguished from successful assistance. Immediate task success may still matter, but it is no longer sufficient evidence that learning has occurred. The decisive question is whether capabilities demonstrated during human-AI interaction remain available when the level of support changes. This is why delayed explanation, adaptation to novel problems, and performance under reduced assistance are especially important outcomes for the theory. Existing evidence increasingly supports this distinction. Learners can produce stronger essays with AI support while showing no comparable gain in underlying knowledge \cite{Fan2025}. Students can feel less burdened during AI-supported inquiry while producing less sophisticated reasoning \cite{Stadler2024}. More broadly, measurable performance gains under AI support do not necessarily correspond to stronger metacognitive processing, retained understanding, or independent transfer \cite{Bastani2025,Song2026,Li2025ProgrammingProcesses}. At the same time, when AI tutoring is structured to scaffold reasoning steps explicitly rather than simply provide answers, it can support both immediate performance and sustained transfer to novel problems in authentic educational settings \cite{Kestin2025,Makransky2025SenseMaking,DeSimone2025NigeriaTutoring}. Transfer under reduced support therefore completes the mechanism set: delegated agency determines what the learner does, epistemic monitoring and verification determine whether the learner remains engaged in judgment, reconstructive internalization determines whether supported performance becomes retained capability, and transfer shows whether learning has in fact occurred. In this sense, Agentivism is not a theory of AI effectiveness in general. It is a theory of under what conditions human capability grows, and under what conditions it does not, in the course of human-AI interaction.

\section*{How Agentivism differs}

Agentivism is closest in spirit to social cognitive views of human functioning because it treats action, regulation, and perceived control as central to learning rather than secondary to it \cite{Bandura2001}. In particular, Bandura's distinction among direct, proxy, and collective agency provides an important foundation for understanding why learners may rely on external actors to achieve desired outcomes. Agentivism extends this line of thought to human-AI interaction by arguing that delegation to AI is not merely a practical convenience but a constitutive feature of the learning process that must itself be theorized. At the same time, Agentivism departs from social cognitive theory in a decisive way: the proxy is no longer simply another human actor or institution, but a generative system that can produce content, recommend actions, and shape the sequence of cognitive activity while remaining outside the normative boundaries of authorship, responsibility, and educational purpose. Agentivism therefore retains the importance of agency, self-efficacy, and regulation, but makes the allocation of task responsibility between learner and AI a first-order explanatory problem.

Agentivism also intersects with traditions of self-regulation and socially shared regulation of learning, but it changes the object of regulation. In conventional accounts, learners regulate goals, strategies, monitoring, and adaptation within their own activity, and in socially shared regulation they co-regulate these processes with others in collaborative settings \cite{Jarvela2023}. These insights remain fundamental, and recent work on hybrid human-AI regulation has already begun to show that learners may regulate with AI, around AI, and sometimes against AI \cite{LanZhou2025,Yan2024,Cukurova2026,Molenaar2022,Ng2024SRLbotScience,Liu2025SDLChatPython,Gyasi2025HumanAICollaboration}. Agentivism builds on this emerging line of work by making a stronger theoretical claim: under conditions of intelligent delegation, regulation is no longer directed only toward one's own cognition or the coordination of human partners, but also toward the boundaries of delegation itself. Learners must regulate what to offload, what to inspect, what to retain responsibility for, and what must later be reconstructed independently. In this sense, Agentivism is aligned with hybrid regulation perspectives but is not reducible to them. Its distinctive contribution is to define learning itself in relation to how delegated performance is converted back into durable human capability.

Relative to constructivism, Agentivism retains the importance of dialogue, interpretation, and participation, but it rejects the assumption that conversational engagement is sufficient evidence of epistemically productive learning. Constructivist traditions are indispensable for explaining why learners develop understanding through interaction, why meaning must be actively made rather than passively received, and why identity and participation matter in socially organized practice \cite{Dewey1938,Piaget1952,Vygotsky1978}. Agentivism accepts all of these points, yet argues that generative AI introduces a new complication: a system can occupy the interactional role of a seemingly knowledgeable interlocutor without satisfying the epistemic conditions that normally justify such a role. A learner may be highly engaged in dialogue with AI and still rely on suggestions that are weakly warranted, culturally averaged, or insufficiently examined. Agentivism therefore preserves constructivism's concern with interaction while adding a stronger account of epistemic monitoring, verification, and reconstructive internalization. Dialogue matters, but under AI conditions, justified learning depends not only on participation in interaction but on how learners evaluate and transform what the interaction produces.

Relative to connectivism, Agentivism also accepts that knowledge is distributed across people, tools, and networks, but it argues that distribution alone no longer provides an adequate account of learning when some nodes become generative, persuasive, and partially agentic \cite{Siemens2005}. Connectivism remains powerful for explaining why learning depends on access to external resources, why knowing where knowledge resides matters, and why network navigation is itself a competence. Agentivism keeps these insights, yet claims that generative AI alters the structure of the learning problem by changing what networked systems do. They do not merely store or route information; they generate candidate explanations, reorganize possibilities, prioritize options, and shape what becomes cognitively salient in the moment of task performance. For this reason, Agentivism is not simply a network theory updated for AI. It is a theory of learning under conditions where networked support can also substitute for parts of reasoning and production, making it necessary to explain how learners preserve judgment and reconstruct assisted performance into retained capability.

Relative to cognitivism, Agentivism keeps mental representation, memory, attention, and cognitive load at the centre of analysis, but it insists that these processes must now be interpreted in relation to delegation and reconstruction. Cognitivism explains why offloading can reduce mental effort, why schemas matter for transfer, and why internal representation remains essential for independent performance \cite{Miller1956,AtkinsonShiffrin1968}. Agentivism agrees, but adds that in human-AI interaction the critical question is no longer only how learners process information internally, but which parts of processing are displaced onto AI and under what conditions the learner later regains functional command of them. Similarly, relative to behaviourism, Agentivism preserves the importance of contingencies and reinforcement while asking a deeper question than whether AI-supported behaviour is strengthened \cite{Watson1913,Skinner1938}. The stronger question is whether the behaviour that changes belongs to the learner in a durable and transferable way, or whether the apparent competence resides mainly in the human-AI configuration at the moment of support. Agentivism therefore reorganizes, rather than replaces, earlier theories. Its distinctive claim is that in the age of generative and agentic AI, learning is best understood as the growth of durable human capability through the selective delegation, epistemic monitoring and verification, and reconstructive internalization of intelligent assistance.

\section*{Empirically testable propositions}

\begin{figure}
    \centering
    \includegraphics[width=1\linewidth]{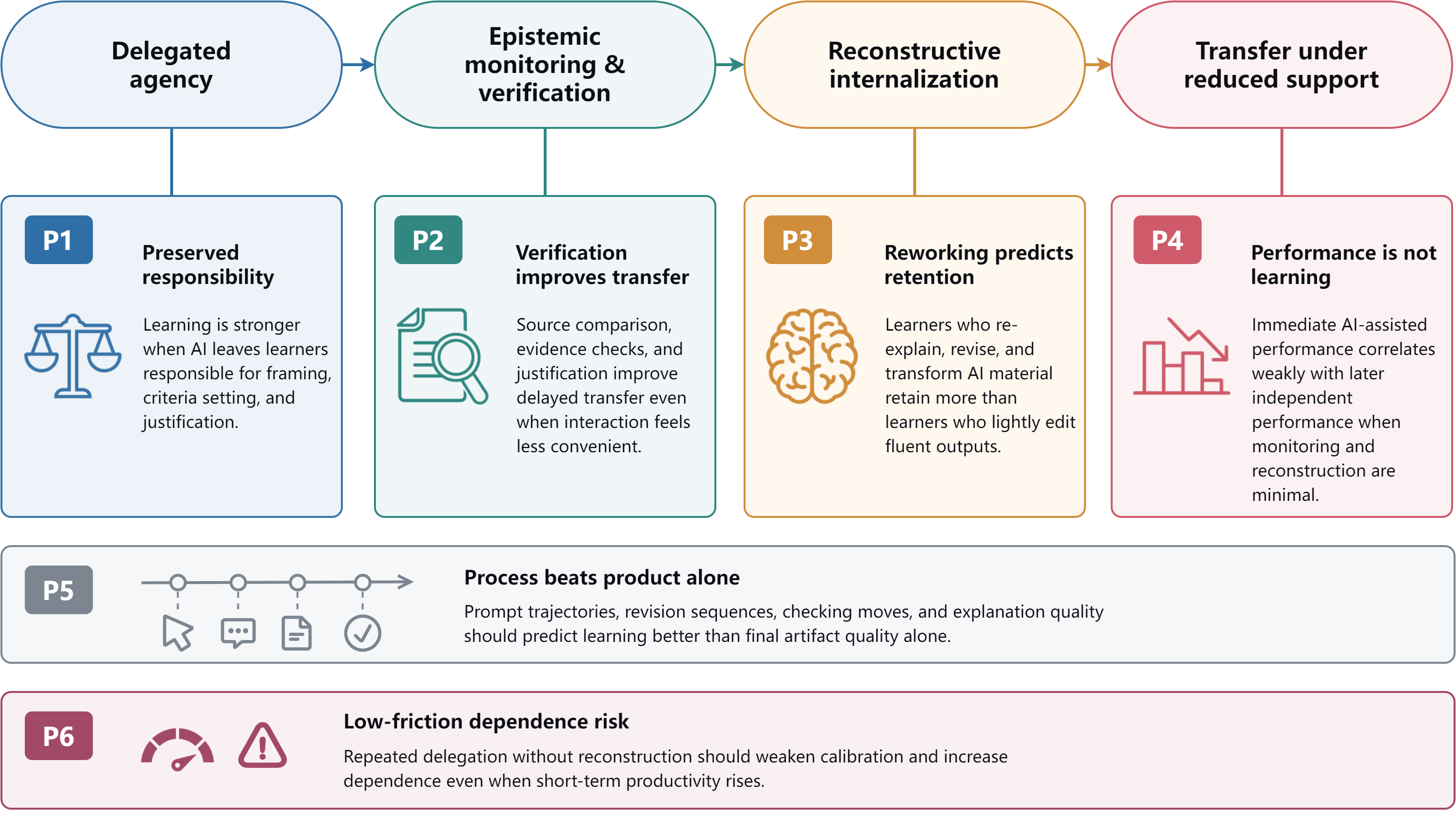}
    \caption{\textbf{Empirical propositions derived from the mechanisms of Agentivism.} This figure maps six empirically testable propositions (P1–P6) onto the four core mechanisms of Agentivism, illustrating how predictions about learning outcomes arise from specific process components.}
    \label{fig:propositions}
\end{figure}

As a mid-range conceptual theory, Agentivism is intended not only to synthesize prior traditions but also to generate testable propositions about how learning unfolds during human-AI interaction (Fig.~\ref{fig:propositions}). Its claims are not that AI support is inherently beneficial or harmful, nor that learners should always minimize delegation. Rather, the theory predicts that learning outcomes will depend on how delegation is structured, monitored, and converted back into retained capability. At minimum, Agentivism implies that studies of learning with AI should move beyond binary comparisons between AI use and non-use and instead examine the processes by which learners allocate agency, verify AI contributions, reconstruct accepted outputs, and perform under reduced support.

A \textit{first proposition} is that learning should be stronger when AI support preserves learner responsibility for problem framing, criteria setting, and justification than when AI support collapses these processes into direct answer delivery. This follows from the mechanism of delegated agency: if AI takes over those functions, the learner has fewer opportunities to practice the cognitive and epistemic operations that later support independent performance. A \textit{second proposition} is that interaction designs requiring verification, comparison with sources, or justification of AI uptake should improve delayed transfer even when they reduce convenience or subjective fluency. This follows from the mechanism of epistemic monitoring and verification and is consistent with evidence that cognitive forcing interventions can reduce over-reliance on AI \cite{Bucinca2021}. A \textit{third proposition} is that learners who substantially re-explain, revise, or transform AI-generated material should show stronger retained understanding than learners who mainly accept or lightly edit fluent outputs. This follows from the mechanism of reconstructive internalization and is broadly aligned with emerging findings on differential uptake of AI-generated suggestions in writing and inquiry \cite{Kim2026,Zheng2025}.

A \textit{fourth proposition} is that immediate AI-assisted performance should correlate only weakly with later independent performance when epistemic monitoring and reconstruction are minimal. This prediction follows directly from the distinction between assisted performance and learning and is already suggested by studies in which stronger AI-supported products do not correspond to stronger underlying knowledge, reasoning, or transfer \cite{Fan2025,Bastani2025,Liu2025,Li2025ProgrammingProcesses}. Conversely, when AI systems are structured to require monitoring, reconstruction, and explicit reasoning steps before moving forward, it can support both immediate performance and sustained transfer to novel problems \cite{Kestin2025,Makransky2025SenseMaking}. A \textit{fifth proposition} is that process measures taken during human-AI interaction should predict later learning better than final product quality alone. Relevant indicators may include prompt trajectories, revision sequences, evidence-checking moves, explanation quality, and the extent to which learners transform rather than merely adopt AI output \cite{Pozdniakov2026,Qian2026,Singh2024}. This follows from the theory's claim that learning is not located solely in the final artifact, but in the sequence through which delegated performance is evaluated and reconstructed. A \textit{sixth proposition} is that repeated low-friction delegation without subsequent reconstruction should be associated over time with weaker calibration of one's own competence and greater dependence on external support, even when short-term productivity rises \cite{MesseriCrockett2024,Clark2025,Rossi2026,Wei2025DigitalStoryGAI}.

These propositions also clarify the level at which Agentivism operates. The theory is not yet a formal computational model of learning, nor does it specify a single universal sequence that all learners must follow in identical form. Instead, it offers a mechanistic explanatory framework that identifies what should vary meaningfully across tasks, designs, and learner populations: the distribution of agency, the quality of verification, the depth of reconstruction, and the endurance of capability after support changes. In this sense, Agentivism is comparable to other influential learning frameworks that explain mechanisms and generate families of hypotheses without reducing learning to a single metric. Its value lies in making a distinctive class of questions empirically visible: not simply whether AI helps learners perform, but when AI-supported performance becomes durable human learning.

\section*{Implications for research and practice}

Agentivism has an immediate implication for research: studies of learning with AI should stop treating ``AI use'' as a single treatment condition \cite{Weidlich2025chatgpt}. What matters is the interactional arrangement through which AI enters the task: whether it gives answers, offers hints, critiques drafts, retrieves evidence, proposes alternatives, or supports multi-step reasoning. From the standpoint of Agentivism, these are not superficial implementation differences because they alter how agency is distributed, how much verification is required, and whether learners are likely to reconstruct supported performance into retained capability. Research should therefore measure not only immediate task outcomes but also the processes through which learners delegate, monitor, revise, and later perform under reduced support. Delayed explanation, adaptation to novel tasks, and independent performance should become more central outcomes, because immediate fluency or correctness can no longer be assumed to index learning \cite{Fan2025,Stadler2024,Darvishi2024,Li2025ProgrammingProcesses}. Evidence from controlled educational studies also indicates that proactive AI scaffolding can improve targeted conceptual learning when support structures are explicit \cite{Yan2025VLA,Chen2025UAVReflectiveScaffold,Makransky2025SenseMaking,DeSimone2025NigeriaTutoring,Ng2024SRLbotScience}.

Agentivism also has a practical implication for pedagogy and assessment: productive use of AI depends less on whether AI is present than on whether the learning design preserves the learner's responsibility for judgment. Instructors should therefore prefer tasks and supports that require learners to frame the problem, articulate criteria, compare AI suggestions with evidence, and explain why accepted outputs are appropriate. In writing, inquiry, and problem solving, the educational aim should not be to eliminate assistance but to ensure that assistance does not replace the learner's role as author, evaluator, and explainer. This principle holds across modalities: well-designed AI tutoring that scaffolds reasoning sequentially, AI-enhanced simulations that require problem decomposition, and AI-supported writing that requires source comparison all show learning gains when the learner remains responsible for judgment and integration \cite{Kestin2025,Lim2025Simulation,Kim2026,Ngu2025Game,Yan2025VLA,Chen2025UAVReflectiveScaffold,Makransky2025SenseMaking}. For assessment, this means that final products are increasingly insufficient indicators of learning. When AI can contribute directly to drafting, solving, or revising, valid assessment requires evidence of the process through which the learner engaged that support, including revision patterns, justification, source checking, and what can later be reproduced or adapted with less assistance \cite{Jiang2026,Pozdniakov2026,Molenaar2022,Singh2024}. Trace-based evidence is therefore important not for surveillance as such, but because it helps recover the distinction between assisted performance and durable human capability.

Finally, Agentivism implies that broader design and governance questions matter insofar as they shape the conditions under which learning mechanisms can operate well. Interface design, institutional rules, accessibility provisions, and norms for acceptable AI assistance all influence whether learners remain active in delegation, verification, and reconstruction or are instead encouraged toward passive uptake. These contextual conditions are especially important where learners differ in prior knowledge, resources, and vulnerability. Inclusive design, calibrated supports, and clear expectations about acceptable delegation are therefore not external add-ons to learning; they shape whether learners have a genuine opportunity to remain agentic while using AI \cite{Rappa2026,Xia2026,Tagare2025}. At the same time, repeated reliance on the same generative AI systems may narrow the diversity of sources, framings, and questions that learners encounter, which means that preserving epistemic diversity is not only a fairness concern but also a learning concern \cite{Traberg2026,Hao2026,DoshiHauser2024}. In this sense, research, pedagogy, assessment, and governance all converge on the same underlying point: if learning is to remain the growth of human capability, then AI support must be evaluated not only by what it helps produce now, but by what it leaves the learner able to do later.

\section*{Limitations and open questions}

Agentivism is proposed as a learning theory, but it does not yet offer a complete formal model of all learning processes involving AI. Its contribution is more modest and more specific: it provides a mechanistic conceptual framework for explaining when AI-supported performance becomes durable human learning and when it does not. In this respect, Agentivism should be understood as a mid-range theory \cite{Merton1968}. It identifies core mechanisms, clarifies their relationships, and generates empirically testable propositions, but it does not claim that all forms of learning through human-AI interaction follow one fixed sequence or can be reduced to a single explanatory principle. This level of theorizing is a strength insofar as it makes a rapidly changing phenomenon conceptually tractable, but it also means that the theory will require refinement as forms of AI support, learner practices, and institutional uses continue to evolve.

A \textit{first open question} concerns operationalization. The core constructs of Agentivism, delegated agency, epistemic monitoring and verification, reconstructive internalization, and transfer under reduced support, are theoretically distinct, but they are unlikely to be captured equally well by a single method or metric. Future research will need to determine how best to identify these mechanisms across settings such as writing, inquiry, collaborative problem solving, tutoring, and professional learning. Some indicators may be visible in trace data, revision histories, or interaction logs; others may require discourse analysis, process measures, delayed assessments, or mixed-method designs. A central challenge is therefore methodological as well as theoretical: the field needs ways to study learning processes during human-AI interaction without collapsing them into product quality alone. One implication is that stronger measurement models will be needed if Agentivism is to support cumulative empirical work rather than remain a persuasive conceptual vocabulary.

A \textit{second open question} concerns boundary conditions. Agentivism argues that learning depends on how delegation is structured, monitored, and reconstructed, but these processes are likely to vary across learner characteristics, disciplinary tasks, developmental stages, and forms of AI support. What counts as productive delegation in novice writing may differ from productive delegation in advanced programming, scientific inquiry, or professional decision making \cite{Li2025ProgrammingProcesses,Choudhuri2024}. Likewise, learners with different levels of prior knowledge, motivation, or self-regulatory skill may not benefit equally from the same arrangement of AI support. More broadly, the theory must still be tested across contexts in which AI contributes not only suggestions and drafts, but also proactive prompts, adaptive scaffolds, multi-agent evaluations, or more autonomous task execution. For this reason, Agentivism should not yet be read as a settled general theory of all human-AI learning. It is better understood as a framework for identifying the variables that should matter most as such learning environments diversify.

A \textit{third open question} concerns timescale. Much of the current evidence on generative AI and learning comes from relatively short tasks or brief interventions, yet the strongest claims of Agentivism concern the development of durable capability over time. The theory predicts that repeated low-friction delegation without reconstruction may weaken independent capability even when short-term performance appears strong, but this prediction remains more plausible than fully established. Longitudinal work is therefore especially important. Researchers will need to examine not only whether AI support improves immediate outcomes, but whether learners become more or less capable, more or less calibrated, and more or less willing to exercise judgment after extended periods of interaction. This temporal question is crucial because the central concern of Agentivism is not whether AI can help now, but what kinds of learners repeated interaction with AI may gradually produce.

These limitations do not weaken the case for Agentivism; they define its research agenda. The theory is needed precisely because existing categories are no longer sufficient to describe learning under conditions where intelligent delegation is easy, persuasive, and increasingly normalized. What remains open is not whether this condition matters, but how finely its mechanisms can be specified, measured, and compared across contexts. In that sense, Agentivism is not presented as a finished doctrine, but as a necessary conceptual advance: a theory that brings the distinction between assisted performance and durable human learning into sharper focus at the moment it becomes most important.

\section*{Final Remark}

The rise of generative and agentic AI does not invalidate classical learning theories. It reveals a new limit that they do not fully resolve on their own. Behaviourism explains why AI-supported activity can become reinforcing. Cognitivism explains why offloading changes mental processing and can weaken the conditions for transfer. Constructivist traditions explain why interaction, dialogue, and participation remain essential to learning. Connectivism explains why knowledge is increasingly distributed across people and systems. Yet none of these traditions, by itself, fully explains what learning becomes when AI systems can contribute directly to task completion and make successful performance possible without commensurate growth in human capability.

Agentivism is an attempt to name that condition precisely and to explain it as a problem of learning rather than merely of technology use. Its central claim is simple: under conditions of human-AI interaction, learning occurs when learners delegate selectively, verify critically, and reconstruct AI-assisted performance into knowledge and skill that remain available beyond the immediate support. What matters, then, is not only what learners can produce with AI, but what they can later explain, adapt, and transfer with less of it. If generative and agentic AI become a lasting part of how people write, inquire, solve problems, and study, then learning theory must be able to distinguish assisted performance from durable human growth. Agentivism is proposed as one way of making that distinction conceptually explicit, empirically tractable, and educationally consequential.

\bibliography{sn-bibliography}


\section*{Competing interests}
The authors declare no competing interests.

\end{document}